\documentclass{article}
\usepackage{arxiv}
\usepackage{booktabs,makecell, multirow, tabularx}
\usepackage[utf8]{inputenc} 
\usepackage[T1]{fontenc}    
\usepackage{hyperref}       
\usepackage{url}            
\usepackage{booktabs}       
\usepackage{amsfonts}       
\usepackage{nicefrac}       
\usepackage{microtype}      
\usepackage{lipsum}
\usepackage{graphicx}
\usepackage{threeparttable}
\usepackage{amsmath,amsfonts}
\usepackage[caption=false,font=normalsize,labelfont=sf,textfont=sf]{subfig}
\graphicspath{ {./images/} }

\title{GraphGSOcc: Semantic-Geometric Graph Transformer with Dynamic-Static Decoupling for 3D Gaussian Splatting-based Occupancy Prediction}

\author{
 Ke Song \\
  School of Intelligent Systems Engineering \\
  Sun Yat-sen University\\
  Shenzhen, China  528406\\
  \texttt{songk9@mail2.sysu.edu.cn} \\
   \And
 Yunhe Wu  \\
  School of Intelligent Systems Engineering \\
  Sun Yat-sen University\\
  Shenzhen, China  528406\\
  \texttt{wuyh87@mail2.sysu.edu.cn} \\
  \And
 Chunchit Siu \\
  School of Intelligent Systems Engineering \\
  Sun Yat-sen University\\
  Shenzhen, China  528406\\
  \texttt{siucc@mail2.sysu.edu.cn} \\
  \And
 Huiyuan Xiong \\
  School of Intelligent Systems Engineering\\
  Sun Yat-sen University\\
  Shenzhen, China  528406\\
  \texttt{xionghy@mail.sysu.edu.cn} \\
}

\begin{document}
\maketitle
\begin{abstract}
Addressing the task of 3D semantic occupancy prediction for autonomous driving, we tackle two key issues in existing 3D Gaussian Splatting (3DGS) methods: (1) unified feature aggregation neglecting semantic correlations among similar categories and across regions, (2) boundary ambiguities caused by the lack of geometric constraints in MLP iterative optimization and (3) biased issues in dynamic-static object coupling optimization. We propose the GraphGSOcc model, a novel framework that combines semantic and geometric graph Transformer and decouples dynamic-static objects optimization for 3D Gaussian Splatting-based Occupancy Prediction. We propose the Dual Gaussians Graph Attenntion, which dynamically constructs dual graph structures: a geometric graph adaptively calculating KNN search radii based on Gaussian poses, enabling large-scale Gaussians to aggregate features from broader neighborhoods while compact Gaussians focus on local geometric consistency; a semantic graph retaining top-M highly correlated nodes via cosine similarity to explicitly encode semantic relationships within and across instances. Coupled with the Multi-scale Graph Attention framework, fine-grained attention at lower layers optimizes boundary details, while coarse-grained attention at higher layers models object-level topology. On the other hand, we decouple dynamic and static objects by leveraging semantic probability distributions and design a Dynamic-Static Decoupled Gaussian Attention mechanism to optimize the prediction performance for both dynamic objects and static scenes.  GraphGSOcc achieves state-ofthe-art performance on the  SurroundOcc-nuScenes, Occ3D-nuScenes, OpenOcc and KITTI occupancy benchmarks. Experiments on the SurroundOcc dataset achieve an mIoU of 25.20\%, reducing GPU memory to 6.8 GB, demonstrating a 1.97\% mIoU improvement and 13.7\% memory reduction compared to GaussianWorld.
\end{abstract}


\section{Introduction}

Recent years have witnessed a significant paradigm shift in autonomous driving perception, transitioning from multi-modal fusion reliant on LiDAR toward vision-centric approaches based predominantly on image data. This evolution is primarily motivated by the potential to reduce dependence on costly LiDAR sensors \cite{xiong2022road}. Within this vision-centric framework, occupancy networks have become foundational, capturing dense 3D structures of the environment \cite{liang2024suprnet}. This burgeoning perception paradigm focuses on inferring the occupancy state of individual voxels within a discretized 3D space, representing a departure from methods predicting 3D bounding boxes \cite{zhu2024drop,liu2022petr,wang2023exploring, 10123008} or vectorized map elements \cite{chen2024maptracker,li2022hdmapnet,liao2022maptr,liao2024maptrv2,yuan2024streammapnet}. Occupancy networks demonstrate strong generalization capabilities, effectively handling open-set objects, irregularly shaped vehicles, and specialized road infrastructure \cite{wang2024panoocc, 10734407, 10745562}.

\begin{figure}
	\centering 
	\includegraphics[width=0.5\linewidth]{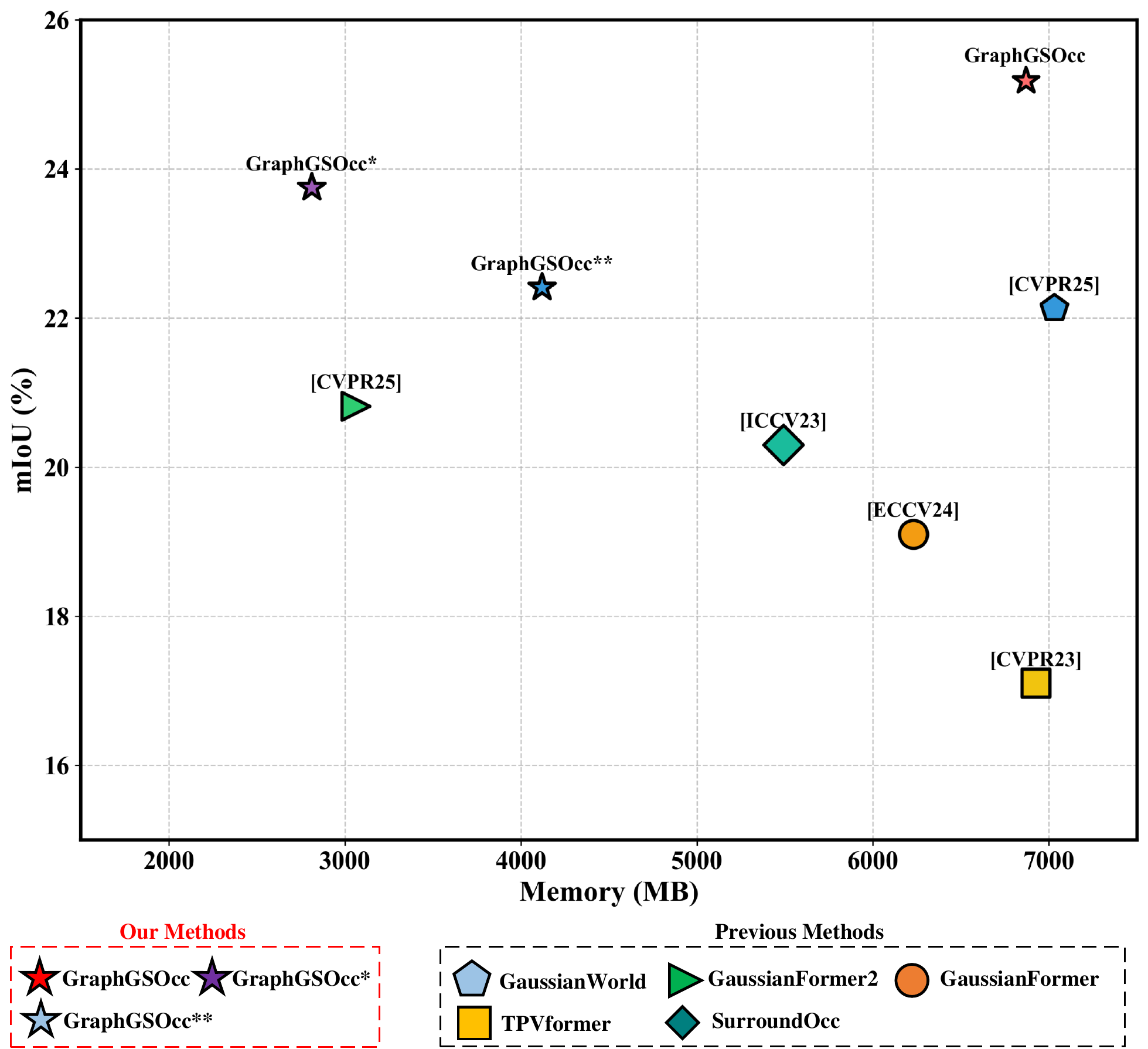}
	\caption{Demonstrating state-of-the-art performance with higher occupancy prediction accuracy and lower computational costs, i.e., memory usage. Data: SurroundOcc \cite{wei2023surroundocc}, GPU: NVIDIA RTX4090.}
	\label{intr}
\end{figure}

3D semantic occupancy prediction, as a dense volumetric segmentation task \cite{cao2022monoscene,tian2023occ3d}, fundamentally requires efficient yet expressive scene representations. While voxel-based approaches \cite{li2023voxformer,wei2023surroundocc} employ dense cubic partitions to capture geometric details with high fidelity, their computational demands scale cubically with spatial resolution due to inherent redundancy in empty space representation. To address this limitation, planar projection techniques such as bird's-eye-view (BEV) \cite{li2024bevformer,yu2023flashocc} and trilinear positional value (TPV) \cite{huang2023tri} have been proposed, which collapse volumetric grids along specific axes to form compact 2D feature maps. However, these dimensionality reduction strategies inevitably retain unnecessary spatial regions in their feature encoding, thereby constraining model efficiency and representational capacity. In contrast to these spatially continuous paradigms, recent object-centric frameworks based on 3D Gaussian Splatting (3DGS) \cite{huang2024gaussianformer} introduce a paradigm shift through probabilistic spatial decomposition, where each Gaussian component parametrizes local geometry (via mean and covariance), visibility (opacity), and semantic attributes in a learnable, sparse formulation. This statistical representation effectively decouples scene complexity from computational cost while preserving semantic expressiveness. However, several limitations persist in the current 3DGS-based methods: (1) Unified feature aggregation neglecting semantic correlations among similar categories and across regions, leading to contextual fragmentation. (2) Iterative refinement via MLPs lacks explicit geometric constraints, causing position drift and semantic ambiguities at object boundaries. (3) Biased issues in dynamic-static object coupling optimization.

To address these challenges, we propose GraphGSOcc, a dynamic structure-aware model that injects geometric and semantic priors into Gaussian interactions. Specifically, we dynamically construct dual graph structures. We implement adaptive radius calculation for KNN search in the geometric graph, where the search range for each Gaussian is dynamically scaled based on its poses, enabling large-scale Gaussians (e.g., road surfaces) to aggregate features from broader neighborhoods while compact Gaussians (e.g., pedestrians) focus on local geometric consistency. For the semantic graph, we compute feature affinity matrices across all Gaussians and selectively retain top-M nodes with highest cosine similarity, explicitly encoding cross-instance and similar instance relationships. Further, a multi-scale aggregation framework hierarchically refines Gaussians—fine-grained attention at lower layers optimizes covariance matrices for boundary details (e.g., small objects prediction), while coarse-grained attention at higher layers models object-level topology (e.g., pedestrian-vehicle motion patterns). Finally, we decouple dynamic and static objects by leveraging semantic probability distributions and design a Dynamic-Static Decoupled Gaussian Attention mechanism to optimize the prediction performance for both dynamic objects and static scenes.

The contributions of our paper are summarized as follows:
\begin{itemize}
	\item[$\bullet$] We propose GraphGSOcc for 3D semantic occupancy prediction which is based on 3DGS.
	\item[$\bullet$] We propose a Dual Gaussian Graph Attention (DGGA) mechanism to dynamically construct geometric and semantic graphs. Additionally, we propose a Multi-scale Graph Attention (MGA) framework to hierarchically refine Gaussians.
	\item[$\bullet$] We propose a Dynamic-Static Decoupled Gaussian Attention (DSDGA) mechanism to optimize the prediction for dynamic and static objects.
	\item[$\bullet$] Our method achieves a mIoU of 25.20\% on Surroundocc \cite{wei2023surroundocc}, while also reducing the GPU memory usage to 6.8 GB, as shown in Fig. \ref{intr}. Our GraphGSOcc also achieves state-of-the-art performance on OpenOcc \cite{graham2015sparse}, Occ3D \cite{tian2023occ3d} and SSCBench-KITTI-360 \cite{geiger2012we}.
\end{itemize}

\begin{figure*}[h]
	\centering
	\includegraphics[width=1.0\textwidth]{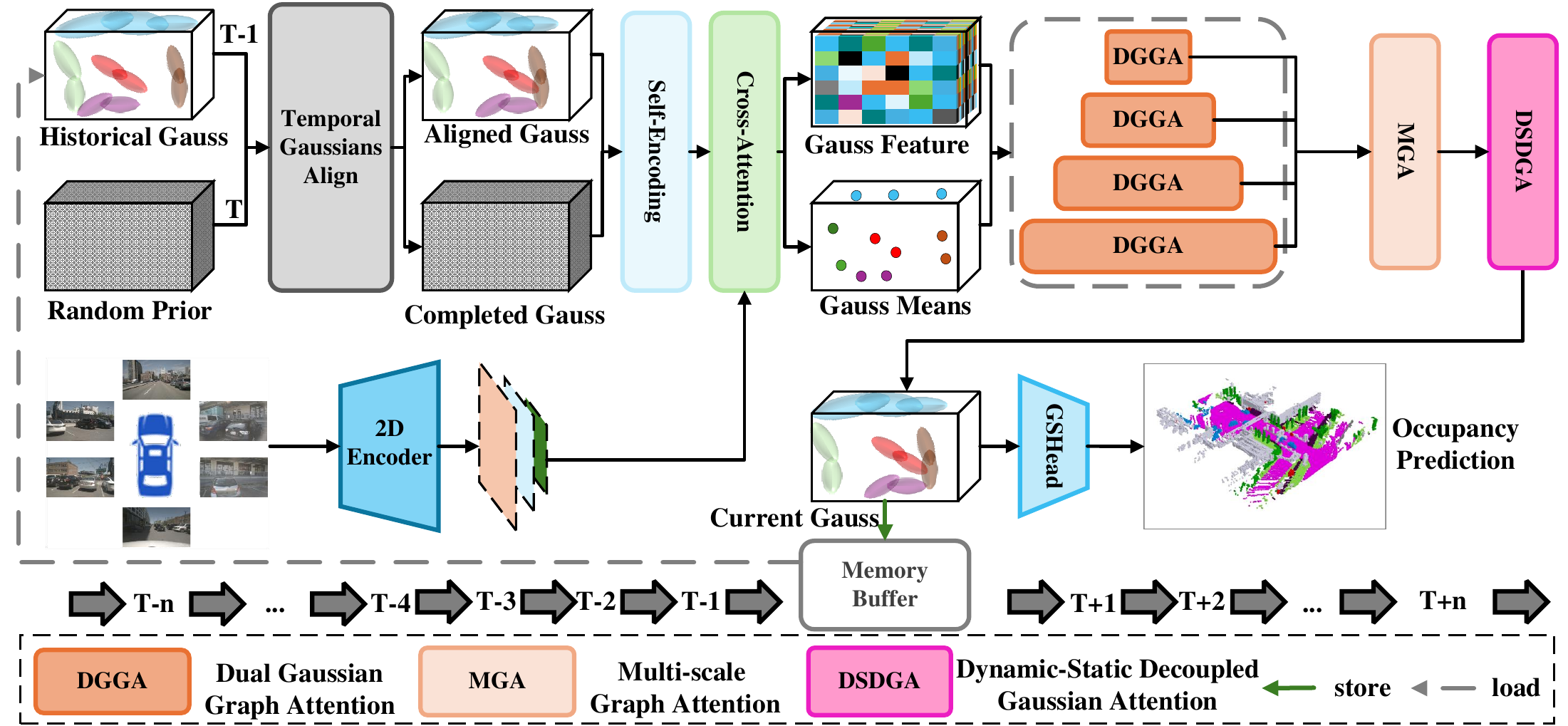}
	\caption{An overview of GraphGSOcc, an efficient yet effective 3D semantic occupancy prediction model that utilizes sequentially incoming multi-view images. GraphGSOcc comprises two key modules:(i) Dual Gaussians Graph Attenntion (DGGA), which dynamically constructs dual graph structures and fuses features from both geometric and semantic graphs; (ii) Multi-scale Graph Attention (MGA), which hierarchically refines Gaussians by fine-grained attention at lower layers and coarse-grained attention at higher layers.}
	\label{fig_overall_framework}
\end{figure*}

\section{Related Work}
\subsection{3D Semantic Occupancy Prediction}
3D semantic occupancy prediction has emerged as a crucial task for understanding surrounding environments by jointly modeling0 geometric and semantic information from visual inputs. Early approaches like MonoScene \cite{cao2022monoscene} pioneered 3D semantic scene reconstruction (SSC) from single RGB images, while subsequent works explored more advanced architectures. Transformer-based methods have shown particular promise, with VoxFormer \cite{li2023voxformer} introducing a two-stage framework for voxel occupancy prediction and OccFormer \cite{zhang2023occformer} proposing a dual-path transformer network that decomposes 3D processing into local and global transformations. Recent innovations have focused on enhancing geometric representations, as seen in COTR's \cite{ma2024cotr} geometry-aware occupancy encoder and GEOcc's \cite{tan2024geocc} integration of implicit and explicit depth modeling. OccGen \cite{wang2024occgen} introduced a diffusion-based approach for multi-modal occupancy refinement, though its voxel-based input impacts computational efficiency. More recently, OccMamba \cite{li2024occmamba} demonstrated the potential of hierarchical Mamba architectures for improving 3D occupancy prediction performance. Although these supervised approaches have made significant contributions, their reliance on detailed voxel-level annotations poses a major challenge. Such annotations are not only expensive to acquire but also limit the scalability of these methods.

\subsection{Scene-based 3D Semantic Occupancy Prediction}
In recent advancements of scene perception for autonomous driving, plane representations have emerged as a competitive approach, presenting compelling alternatives to conventional LiDAR-based methods. Among these, BEVFormer \cite{li2024bevformer} was the first to utilize purely camera inputs while achieving comparable performance in detection and segmentation tasks. This method transforms image features into bird's-eye-view (BEV) \cite{huang2021bevdet,li2023bevstereo,li2023bevdepth,liu2023bevfusion,jiang2023polarformer} features as a unified scene representation, leveraging the viewpoint's information diversity for downstream applications. Nevertheless, the BEV paradigm encounters constraints in 3D occupancy prediction because of its inherent loss of height information \cite{wei2023surroundocc}. To address this, TPVFormer \cite{huang2023tri} introduced an enhanced tri-perspective view representation that explicitly integrates height dimensions, better suiting 3D scene understanding. Concurrently, parallel developments have explored voxel-based representations \cite{li2023voxformer,wei2023surroundocc,tan2025geocc,ma2024cotr,wang2024not} as a more precise option for 3D volumetric semantic prediction, providing finer granularity. However, these grid-based approaches uniformly handle all voxels without considering environmental sparsity, leading to inherent inefficiencies in representation.

\subsection{3D Gaussian Splatting for 3D Occupancy Prediction}
The field of 3D scene representation has seen significant advances through Gaussian-based approaches, offering improved efficiency and flexibility over traditional voxel-based methods like Neural Radiance Fields (NeRF). 3D Gaussian Splatting (3DGS) \cite{kerbl20233d} pioneered this direction by employing learnable Gaussians as compact scene representations, significantly enhancing both training and rendering efficiency. Building on this foundation, GaussianFormer \cite{huang2024gaussianformer} introduced an object-centric paradigm using sparse 3D semantic Gaussians, where each Gaussian flexibly represents a region of interest with associated semantic features. Its subsequent variant GaussianFormer2 \cite{huang2024gaussianformer2} further advanced the probabilistic interpretation of scene geometry by modeling each Gaussian as an occupancy probability distribution and employing probabilistic multiplication for geometric reasoning. GaussianWorld \cite{zuo2025gaussianworld} employ a Gaussian world model to explicitly exploit these priors and infer the scene evolution in the 3D Gaussian space considering the current RGB observation.Practical applications have demonstrated the versatility of this approach, with GaussianOcc \cite{gan2024gaussianocc} achieving notable training efficiency gains through Gaussian splatting-based rendering, while GSRender \cite{sun2024gsrender} specifically addressed the challenge of duplicated predictions along camera rays inherent in 2D-supervised systems. More recently, GaussTR \cite{jiang2024gausstr} has extended these capabilities to open-vocabulary semantic understanding by combining sparse Gaussian queries with pre-trained vision foundation models, eliminating the need for explicit annotations while maintaining strong generalization performance. This progression of Gaussian-based methods continues to push the boundaries of 3D scene representation across efficiency, accuracy, and semantic understanding.

Our proposed GraphGSOcc diverges from prior works in three works: (1) By dynamically constructing the graph structure between Gaussian distributions, the Dual Gaussian Graph Attention (DGGA) enhances local geometric and semantic correlations while preserving global modeling capabilities. (2) A Multi-scale Graph Attention (MGA) module is proposed to enhance the geometric detail prediction of small objects. (3) We decouple dynamic and static objects by leveraging semantic probability distributions and design a Dynamic-Static Decoupled Gaussian Attention mechanism to optimize the prediction performance for both dynamic objects and static scenes. 

\begin{figure*}[h]
	\centering
	\includegraphics[width=1.0\textwidth]{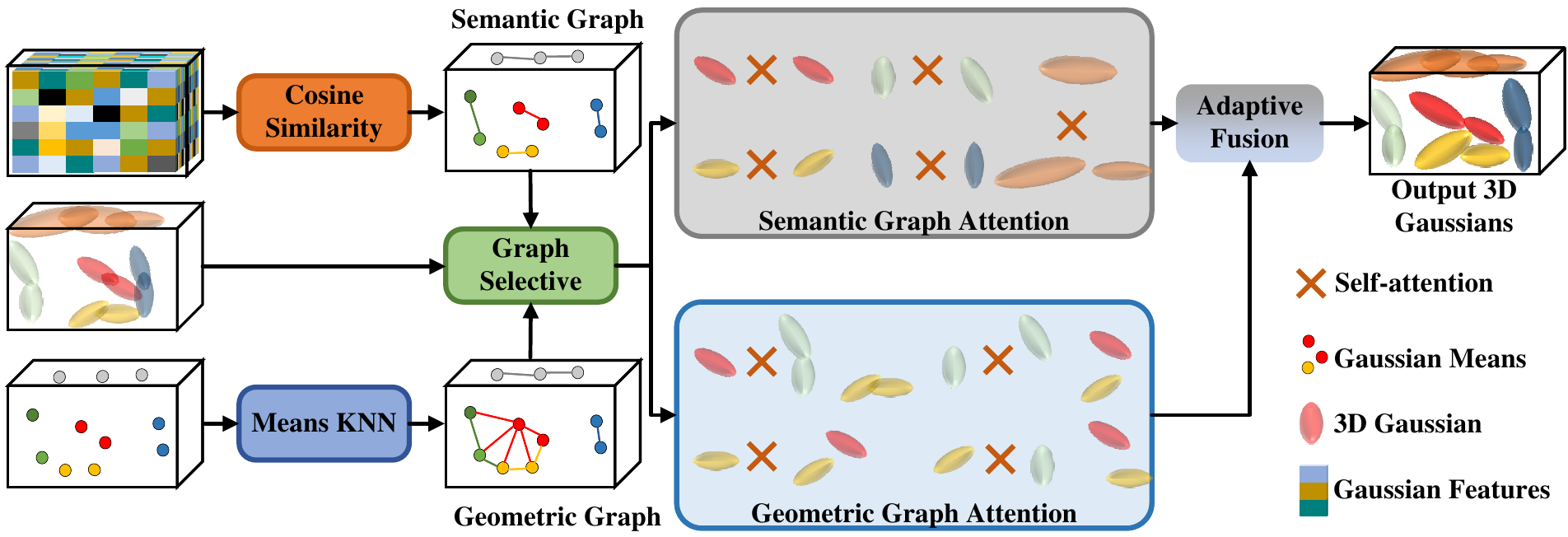}
	\caption{An overview of our Dual Gaussians Graph Attention (DGGA) module, which dynamically constructs dual graph structures and fuses features from both geometric and semantic graphs. DGGA consists of three main steps: (i) Construct geometric graph and semantic graph, (ii) Geometric and Semantic Graph Attention, and (iii) Adaptive Fusion.}
	\label{fig_overall_ddga}
\end{figure*}

\begin{figure}[h]
	\centering
	\includegraphics[width=0.5\linewidth]{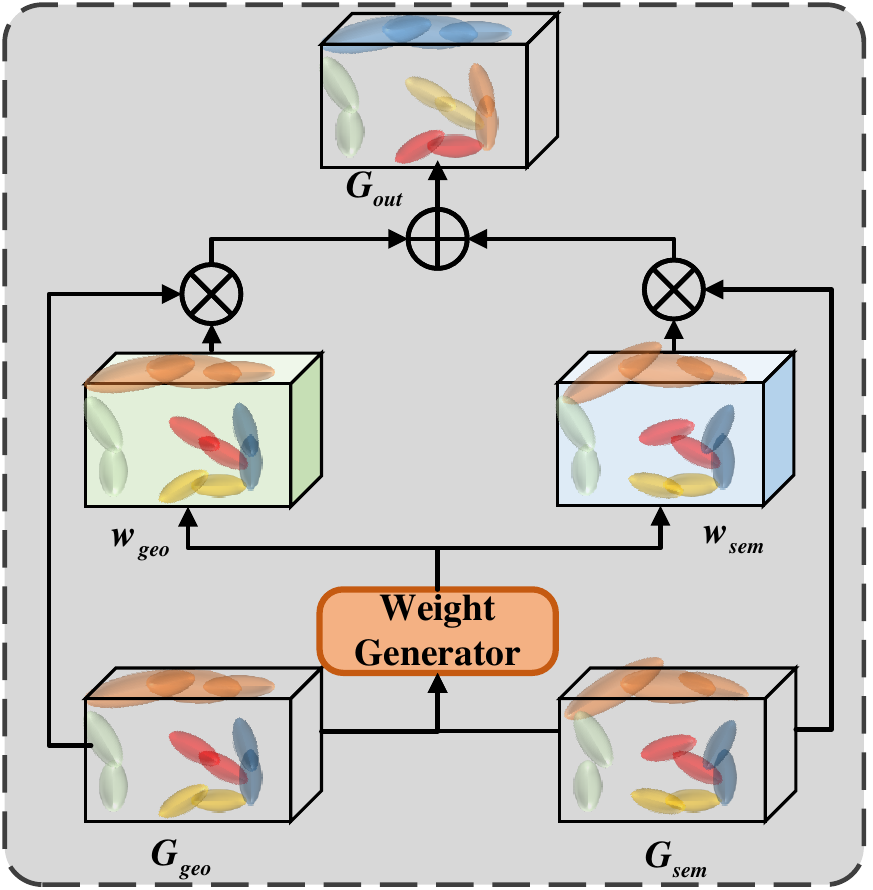}
	\caption{An overview of Adaptive Fusion, which fuses features from both geometric and semantic graphs. The output of Adaptive Fusion is the enhanced 3D Gaussian.}
	\label{fig_af}
\end{figure}

\begin{figure}[h]
	\centering
	\includegraphics[width=0.5\linewidth]{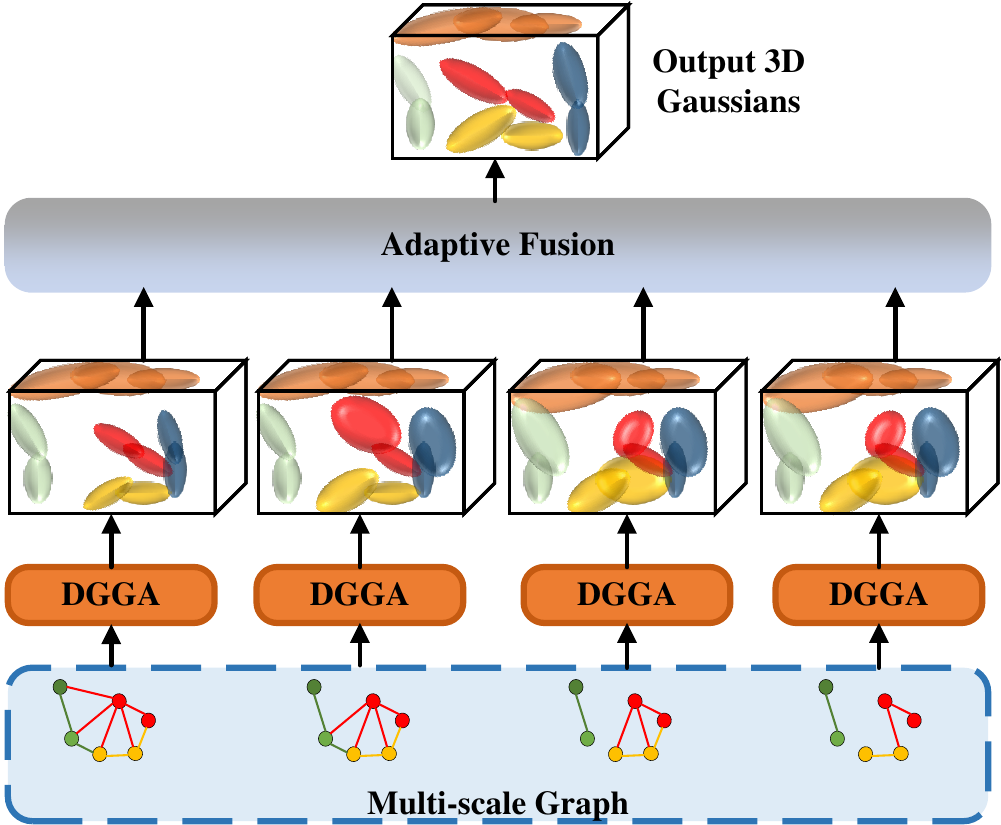}
	\caption{An overview of Multi-scale Graph Attention (MGA) framework, fine-grained attention at lower layers optimizes boundary details, while coarse-grained attention at higher layers models object-level topology.}
	\label{fig_mga}
\end{figure}

\begin{figure*}[h]
	\centering
	\includegraphics[width=\linewidth]{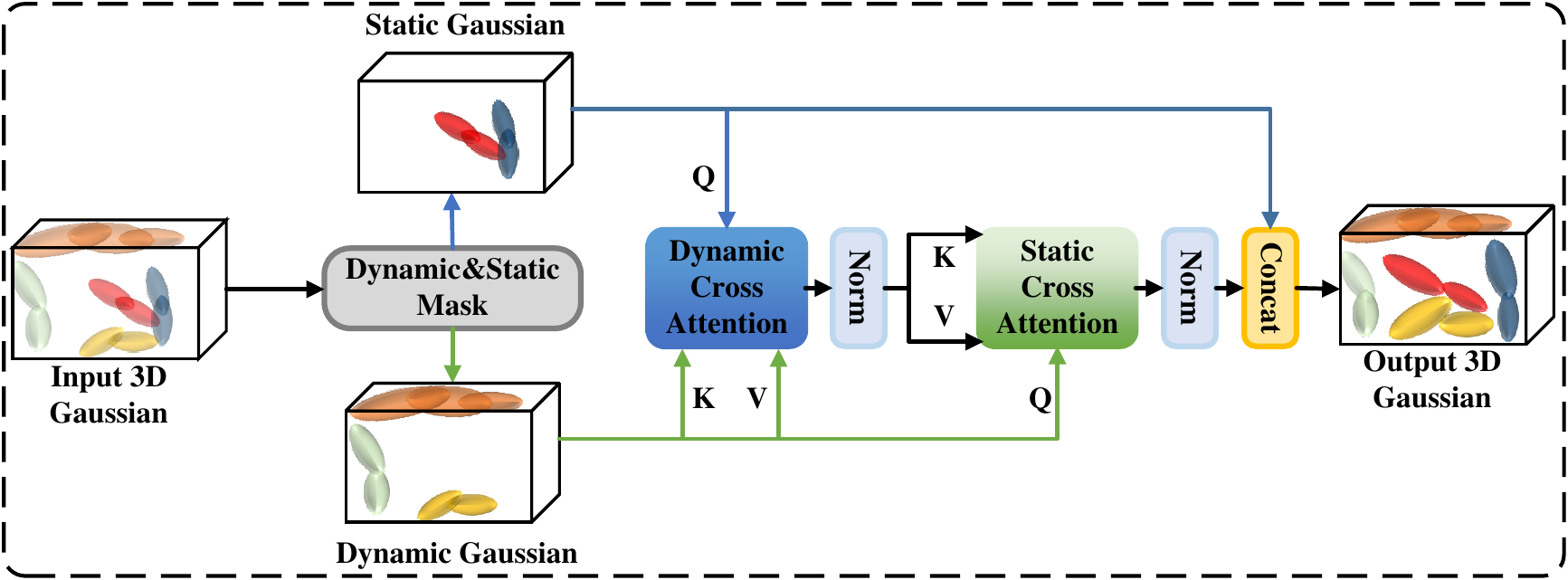}
	\caption{An overview of Dynamic-Static Decoupled Gaussian Attention (DSDGA) framework, }
	\label{fig_dsdga}
\end{figure*}
\section{Proposed Method}
In this section, we outline how the proposed GraphGSOcc is utilized to enhance the performance for 3D semantic occupancy prediction task. In the following sections, we first provide a brief introduction of the overall framework in Section \ref{Framework Overview}.  Thereafter, we provide a detailed explanation of Dual Gaussians Graph Attention (DGGA) which is the core component of GraphGSOcc in Section \ref{DGGA}. Additionally, the Dynamic-Static Decoupled Gaussian Attention (DSDGA) layer, which is the other core component of GraphGSOcc, is demonstrated in details in Section \ref{DSDGA}. Finally, we introduce the other components of GraphGSOcc in Section \ref{other}.

\subsection{Framework Overview}\label{Framework Overview}
Fig. \ref{fig_overall_framework} depicts the pipeline of our GraphGSOcc. As denoted in Fig. \ref{fig_overall_framework}, the input data for GraphGSOcc consists of surround-view images sequence, while the output is 3D semantic occupancy prediction results. At current frame $t$, surround-view images are first processed by \textit{the 2D Encoder}. Then, we load the historical frame $t-1$ gaussians and initialize the random prior gaussians. We use \textit{Temporal Gaussians Align} based on GaussianWorld to get the aligned gaussians and complete gaussians at current frame $t$. Additionally, we use \textit{Self-Encoding} and \textit{Cross-Attention} to optimize the gaussians based on the image features. 
Furthermore, the two key components: (i) \textit{Dual Gaussian Graph Attention (DGGA)} is introduced to enhance and refine the geometric and semantic correlations between gaussians by the means and the feature of gaussians, and (ii) \textit{Dynamic-Static Decoupled Gaussian Attention (DSDGA)} is introduced to decouple the dynamic and static gaussians. Finally, the enhanced gaussians are used to generate the 3D occupancy at current frame $t$ via \textit{GSHead}.

\subsection{Dual Gaussian Graph Attention}\label{DGGA}
In this section, as shown in Figs. \ref{fig_overall_ddga}, we explain (i) how we build the geometric graph structure and semantic graph structure and (ii) how we aggregate the features of neighbor nodes to the center nodes and (iii) how we fuse the features of geometric graph and semantic graph and (iv) how we adaptively fuse the multi-scale gaussian features.

\subsubsection{Construct geometric graph and semantic graph}\label{Geometric Graph and Semantic Graph}
Given the input of 3D Gaussian $G$, Gaussian means $G_m$, and Gaussian features $G_f$, during the construction of the geometric graph structure, we calculate the top-K nearest neighbor nodes for each central node based on the mean of each Gaussian node:
\begin{equation}
	\begin{aligned}
	&inner_{i,j}=-2(G_{m}[i]\cdot G_{m}[j]^T),\\
	&Distance_{i,j} = -{\textstyle \sum_{c=1}^{C}}(G_{m}[i][c])^2 + inner_{i,j} \\
	&- {\textstyle \sum_{c=1}^{C}}(G_{m}[j][c])^2,\\
	&idx_{i}^{geo} = argsort(Distance_{i})[:K]
	\end{aligned}
\end{equation}
where $C$ is the dimension of the Gaussian mean. The $i$ is the central nodes and $j$ is the neighbor nodes. The $K$ is the Top-K. The $idx_{i}^{geo}$ is the index of the neighbor nodes of the central node $i$ based on geometric distance. Following the geometric  indices $idx_{i}^{geo}$, we construct the geometric graph. 

Similarly, during the construction of the semantic graph structure, we calculate the top-M nearest neighbor nodes for each central node based on the feature of each Gaussian node:
\begin{equation}
	\begin{aligned}
	&sim_{i,j}=\frac{G_{f}^{i}\cdot G_{f}^{j}}{||G_{f}^{i}||\cdot ||G_{f}^{j}||}, \\
	&idx_{i}^{sem} = argsort(sim_{i})[:M]
	\end{aligned}	
\end{equation}
where the $M$ is the Top-M. The $idx_{i}^{sem}$ is the index of the neighbor nodes of the central node $i$ based on semantic similarity. Following the semantic indices $idx_{i}^{sem}$, we construct the semantic graph. 

\subsubsection{Geometric and Semantic Graph Attention}\label{Geometric Graph and Semantic Graph Attention}
First, we compute the Geometric Graph Attention (GGA) based on the 3D Gaussian and the geometric index $idx_{i}^{geo}$, and fuse the features of neighbor nodes $j$ into the central nodes $i$ through the attention calculation:
\begin{equation}
	\begin{aligned}
	&Q_{geo},K_{geo},V_{geo} = W_{Q}G, W_{K}G,W_{V}G,\\
	&K_{geo}^{nei},V_{geo}^{nei}=K_{geo}[idx^{geo}],V_{geo}[idx^{geo}],\\
	&G_{geo}=Softmax(\frac{Q_{geo}\cdot (K_{geo}^{nei})^T}{\sqrt{d_{k}}}) \cdot V_{geo}^{nei}
	\end{aligned}	
\end{equation}
where $W$ is the projection layer, $G_{geo}$ is the 3D Gaussian after geometric graph  optimization.

Furthermore, we compute the Semantic Graph Attention (SGA) based on the 3D Gaussian and the geometric index $idx_{i}^{sem}$, and fuse the features of neighbor nodes $j$ into the central nodes $i$ through the attention calculation:
\begin{equation}
	\begin{aligned}
	&Q_{sem},K_{sem},V_{sem} = W_{Q}G, W_{K}G,W_{V}G,\\
	&K_{sem}^{nei},V_{sem}^{nei}=K_{sem}[idx^{sem}],V_{sem}[idx^{sem}],\\
	&G_{sem}=Softmax(\frac{Q_{sem}\cdot (K_{sem}^{nei})^T}{\sqrt{d_{k}}}) \cdot V_{sem}^{nei}
	\end{aligned}	
\end{equation}
where $G_{sem}$ is the 3D Gaussian after semantic graph optimization.

\begin{figure*}
	\centering 
	\includegraphics[width=\textwidth]{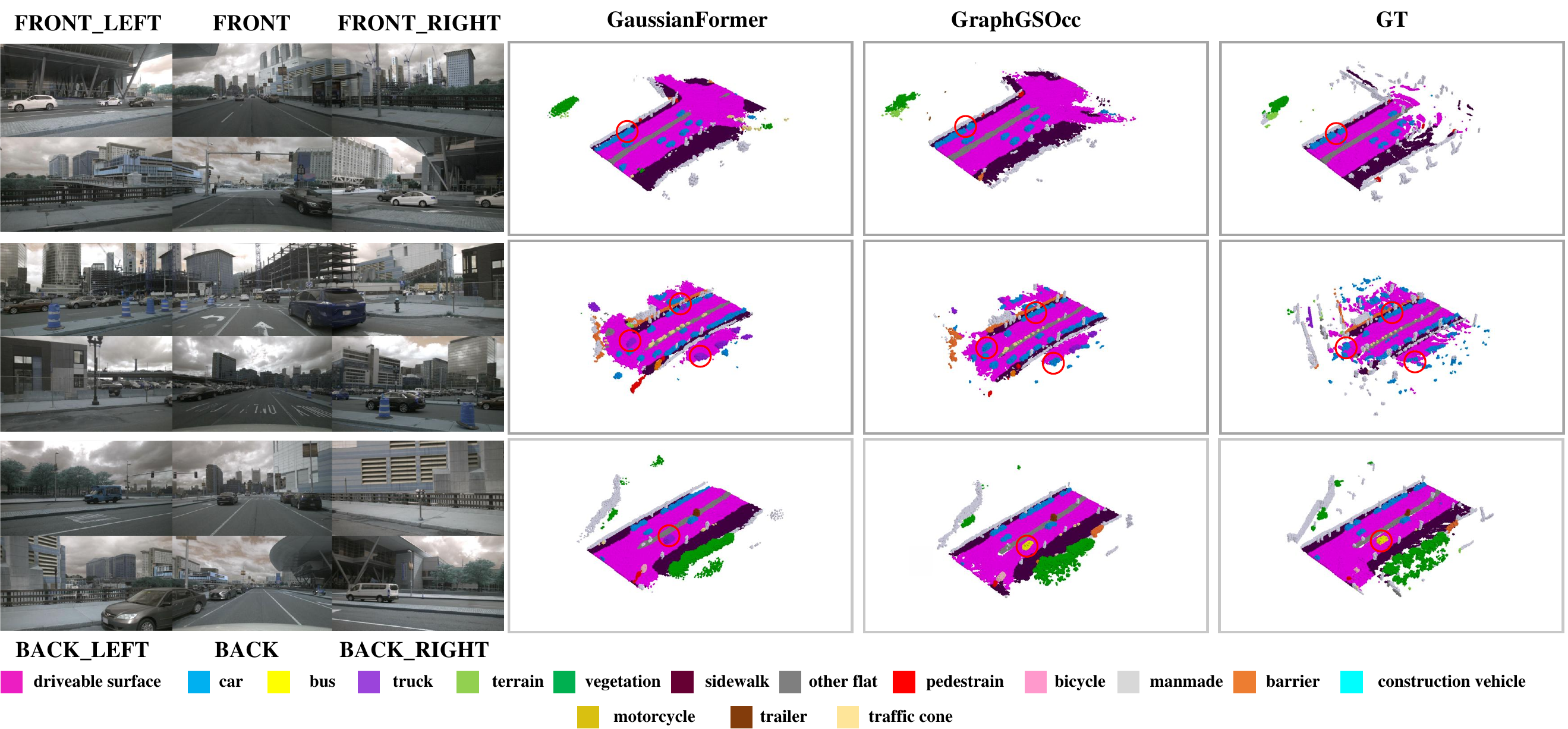}
	\caption{Visual comparison of our GraphGSOcc with state-of-the-art methods on SurroundOcc. Compared to GaussianFormer, our GraphGSOcc generates more precise prediction results of semantic classes and complex scenarios (labeled in red circles).}
	\label{sota}
\end{figure*}
\begin{figure*}
	\centering 
	\includegraphics[width=\textwidth]{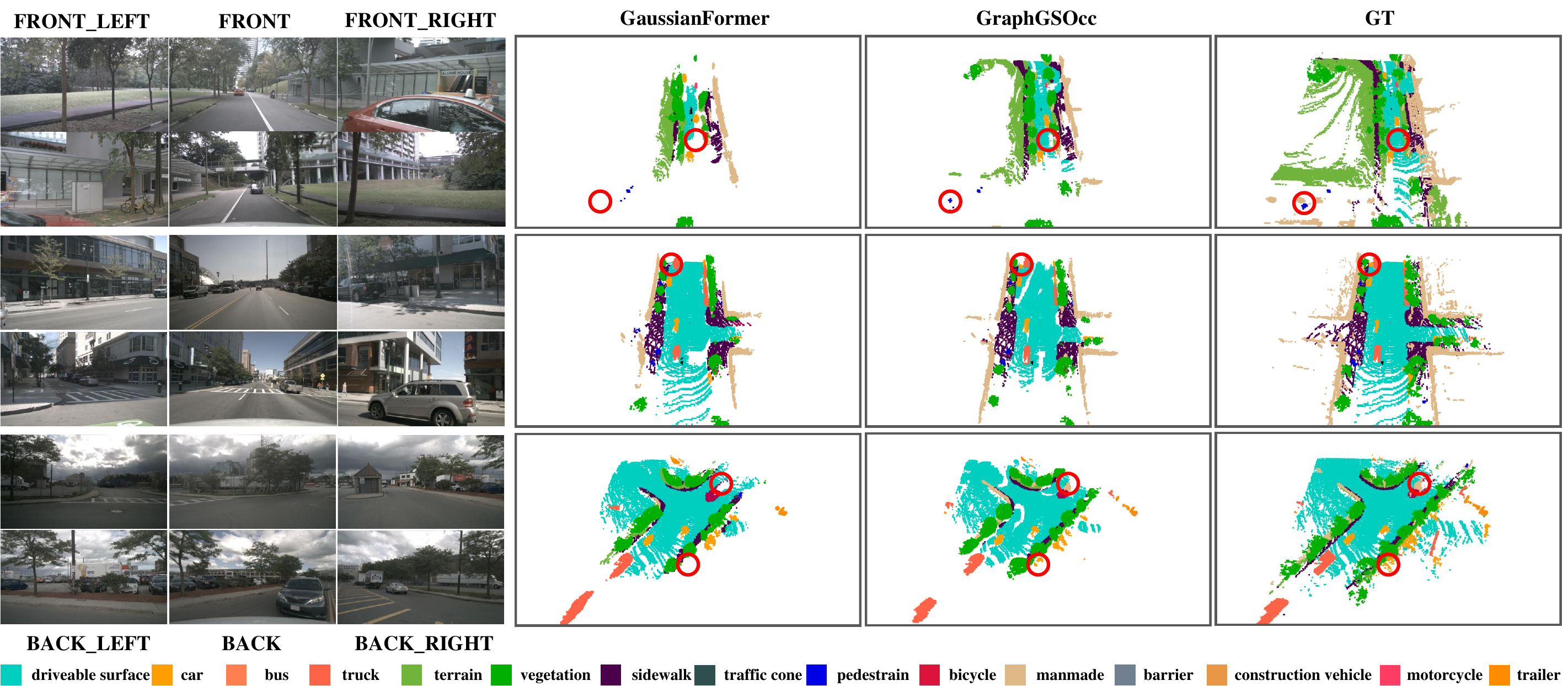}
	\caption{Visual comparison of our GraphGSOcc with state-of-the-art methods on Occ3D. Compared to GaussianFormer, our GraphGSOcc generates more precise prediction results of semantic classes and complex scenarios (labeled in red circles).}
	\label{sota2}
\end{figure*}
\begin{table*}[!t]
	\belowrulesep=0pt
	\aboverulesep=0pt
	\centering     
	\begin{threeparttable}
		\caption{Performance comparison of 3D Occupancy Prediction Methods on the NuScenes SurroundOcc Dataset. The Latency and Memory Consumption for All Methods are Tested on One NVIDIA 4090 GPU with Batchsize One. The Bold and Underline Indicate the Best and Second-best Tesults, respectively. (TPVFormer* is Supervised by Dense Occupancy Annotations. GraphGSOcc**  Denotes Ours Method Based on GaussianFormer. GraphGSOcc*  Denotes Ours Method Based on GaussianFormer2. GraphGSOcc Denotes Our Method Based on GaussianWorld)}\label{tbl1}
		\renewcommand{\arraystretch}{1.5}
		\begin{tabular}{c|c|c|c|c|c|c|c}
			\toprule
			Methods & Venue& Backbone & Input Size & mIoU $\uparrow$ & IoU $\uparrow$ & Latency (ms) $\downarrow$ & Memory (MB) $\downarrow$ \\
			\hline
			MonoScene & CVPR'22 & ResNet-101  & 1600 $\times$ 928  & 7.31 & 23.96 & - & -\\
			BEVFormer & ECCV'22 & ResNet-101 & 1600 $\times$ 900 & 16.75 & 30.50 & \textbf{212} & 6651 \\
			TPVFormer & CVPR'23 & ResNet-101  & 1600 $\times$ 928  & 11.66 & 11.51 & 341 & 6926 \\
			TPVFormer* & CVPR'23 & ResNet-101  & 1600 $\times$ 928  & 17.10 & 30.86 & 341 & 6926 \\
			OccFormer & ICCV'23 & ResNet-101  & 1600 $\times$ 928  & 19.03 & 31.39 & - & - \\
			SurroundOcc & ICCV'23 & ResNet-101 & 1600 $\times$ 900  & 20.30 & 31.49 & 312 & 5491 \\
			
			\hline
			GaussianFormer & ECCV'24 & ResNet-101 & 1600 $\times$ 864 & 19.10 & 29.83 & 372 & 6229 \\
			GraphGSOcc** & - & ResNet-101 & 1600 $\times$ 864 & 22.41 & 34.21 & 280 & \underline{4120} \\
			\hline
			GaussianFormer2 & CVPR'25 & ResNet-101 & 1600 $\times$ 864 & 20.82 & 31.74 & 357 & 3063\\
			GraphGSOcc* & - & ResNet-101 & 1600 $\times$ 864 & \underline{23.75} & \underline{35.58} & \underline{270} & \textbf{2812}\\
			\hline
			GaussianWorld & CVPR'25& ResNet-101&1600 $\times$ 864 & 22.13 & 33.40 & 381 & 7030\\
			GraphGSOcc & - & ResNet-101 & 1600 $\times$ 864 & \textbf{25.20} & \textbf{36.11} & 368 & 6063\\
			\bottomrule
		\end{tabular}
	\end{threeparttable}
\end{table*}

\begin{table}
	\belowrulesep=0pt
	\aboverulesep=0pt
	\centering
	      
	\begin{threeparttable}
		\caption{3D Occupancy Prediction Performence on OpenOcc and Occ3D Dataset. The RayIoU Results are Obtained Using the Annotations from OpenOcc \cite{graham2015sparse}, While the MIoU Results are Based on the Occ3D Annotations \cite{tian2023occ3d}. Bold and Underline Denote the Best Performence and the Second-Best Performence.}\label{tbl_open_occ3d}
		\renewcommand{\arraystretch}{1.5}
		\begin{tabular}{c|cccc}
			\toprule
			Methods   & Backbone& Input Size & $\text{RayIoU}$  &$\text{mIoU}$ \\
			\hline
			RenderOcc  & R101 & 1600$\times$900 & 19.5 & 24.6 \\
			OccNet  & R101 & 1600$\times$900 & 32.7  & 36.8\\
			BEVFormer  & R101 & 1600$\times$900 & 28.1 & 39.1\\
			FB-Occ  & R101 & 1600$\times$900 & 33.5  & 39.4\\
			SparseOcc  & R101 & 1600$\times$900 & 35.1 & 30.6\\
			\hline
			GraphGSOcc & R101 & 1600$\times$900 & \textbf{36.7} & \textbf{42.6}\\
			\bottomrule 
		\end{tabular}
	\end{threeparttable}
\end{table}

\begin{table}[!t]
	\belowrulesep=0pt
	\aboverulesep=0pt
	\centering
	      
	\begin{threeparttable}
		\caption{Results on the SSCBench-KITTI-360 Test Set \cite{geiger2012we} with a Monocular Camera. GraphGSOcc Achieves New State-of-the-art, Achieving Strong Performance in MIoU and IoU}\label{tbl_kitti}
		\renewcommand{\arraystretch}{1.5}
		\begin{tabular}{c|cccc}
			\toprule
			Methods   & Backbone& Input & $\text{IoU}$  &$\text{mIoU}$ \\
			\hline
			MonoScene  & R101 & C & 37.87 & 12.31 \\
			Voxformer  & R101 & C & 18.76  & 11.91\\
			TPVFormer  & R101 & C & 40.22 & 13.64\\
			OccFormer  & R101 & C & 40.27  & 13.87\\
			GaussianFormer  & R101 & C & 35.38 & 12.92\\
			GaussianFormer-2  & R101 & C & 38.37 & 13.90\\
			\hline
			GraphGSOcc & R101 & C & \textbf{40.92} & \textbf{15.58}\\
			\bottomrule 
		\end{tabular}
	\end{threeparttable}
\end{table}

\begin{figure*}
	\centering 
	\includegraphics[width=\linewidth]{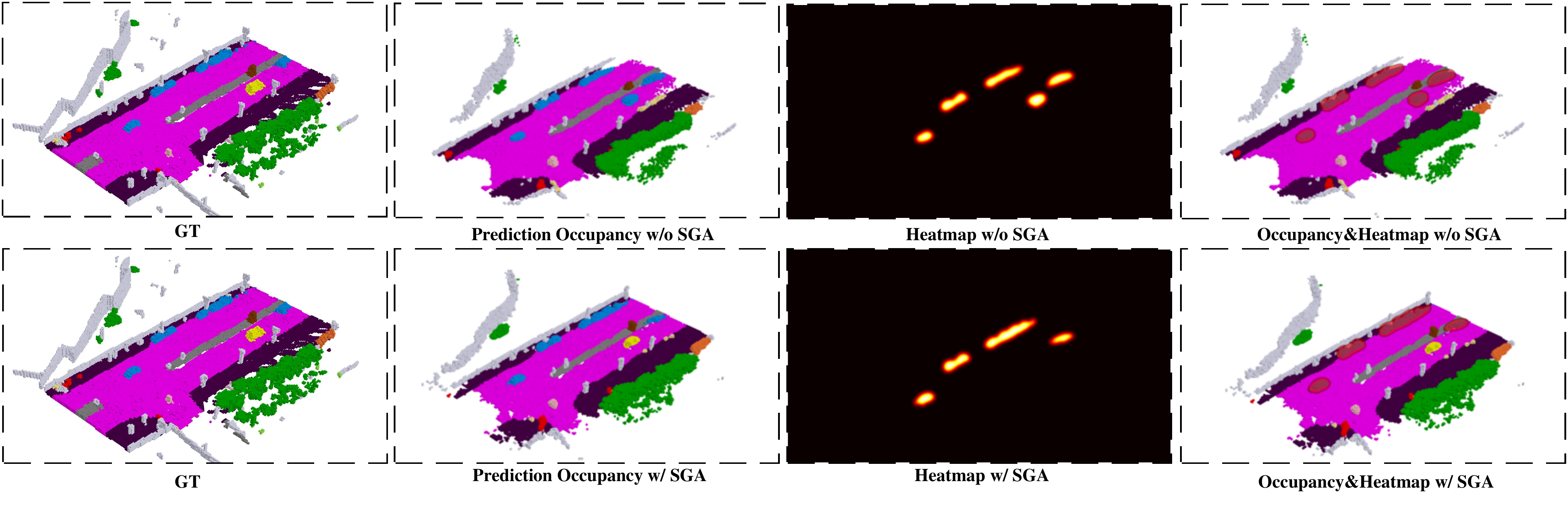}
	\caption{Demonstrating the effect of the SGA in the "car" objects. The first row is the model without SGA module. The second row is the model with SGA module. The first column is the GT. The second column is the prediction occupancy. The third column is the heatmap. The forth column is the occupancy conbined with heatmap.} 
	\label{abla_heatmap}
\end{figure*}

\begin{table}[!t]
	\belowrulesep=0pt
	\aboverulesep=0pt
	\centering     
	\begin{threeparttable}
		\caption{Ablation on the Proposed DGGA Layer without DSDGA. Dyna. Object Denotes Dynamic Object. Stat. Object Denotes Static Object.}\label{tbl_abla_dgga}
		\renewcommand{\arraystretch}{1.5}
		\begin{tabular}{c|ccc|c}
			\toprule
			\multirow{2}{*}{Methods} & \multicolumn{3}{c|}{mIoU} & \multicolumn{1}{c}{IoU} \\ 
			\cline{2-5}
			& All & Dyna. object & Stat. object & All\\ 
			\hline
			baseline & 18.41 & 16.73 & 20.09 & 29.23 \\
			w/ MLP & 19.10 & 16.99 & 21.21 & 29.83 \\
			w/ GGA & 19.83 & 17.21 & 22.45 & 31.98 \\
			w/ SGA  & 20.52 & 18.33 & 22.71 & 30.21 \\
			w/ DGGA  & 21.01 & 18.54 & 23.48 & 32.32 \\
			\bottomrule
		\end{tabular}
	\end{threeparttable}
\end{table}

\begin{table}[!t]
	\belowrulesep=0pt
	\aboverulesep=0pt
	\centering     
	\begin{threeparttable}
		\caption{Ablation on the MGA module without DSDGA.}\label{tbl_abla_mga}
		\renewcommand{\arraystretch}{1.5}
		\begin{tabular}{c|c|c|c}
			\toprule
			Methods & mIoU & IoU & Memory (MB)\\ 
			\hline
			Add & 19.52 & 31.25 & 3080  \\
			Concat & 19.48 & 31.10 & 3288  \\
			Conv Fusion & 20.21 & 31.58 & 3250 \\
			Adaptive Fusion  & 21.31 & 33.41 & 3320 \\
			\bottomrule
		\end{tabular}
	\end{threeparttable}
\end{table}

\begin{figure}
	\centering 
	\includegraphics[width=0.5\linewidth]{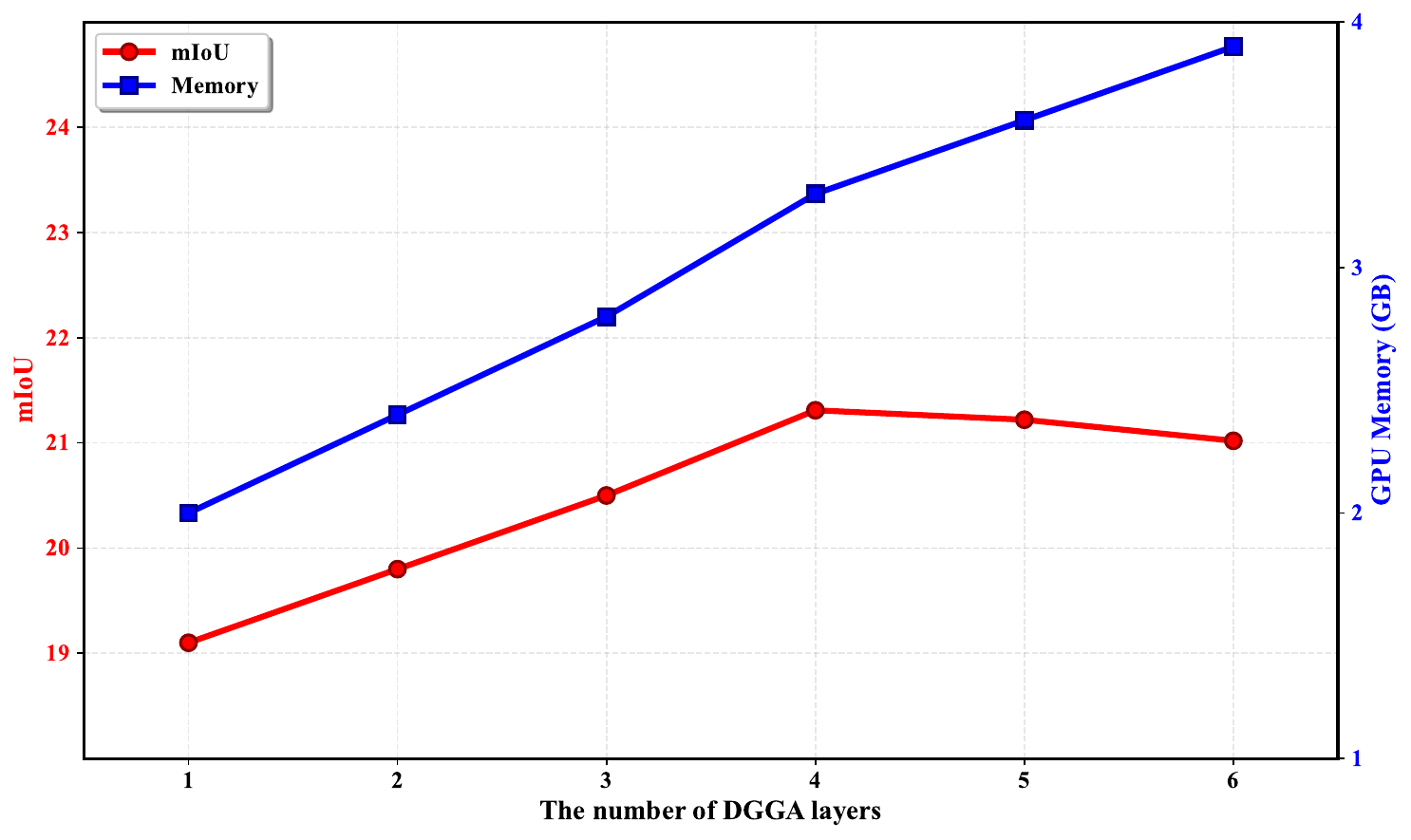}
	\caption{Demonstrating the effect of the numbers of DGGA layer in terms of inference memory and mIoU.}
	\label{abla_number_dgga}
\end{figure}

\begin{figure}
	\centering 
	\includegraphics[width=0.5\linewidth]{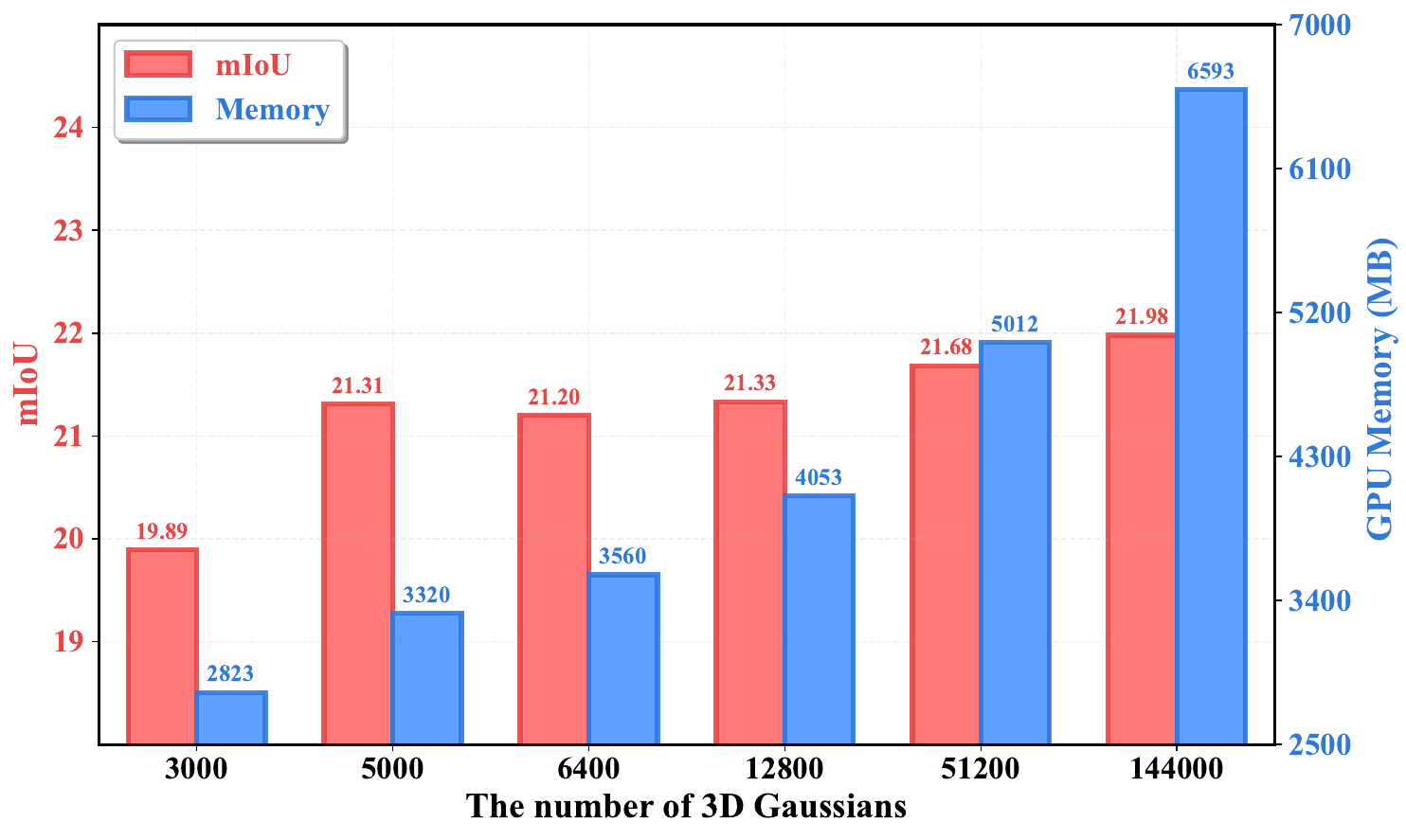}
	\caption{Demonstrating the effect of the numbers of 3D Gaussians in terms of inference memory and mIoU.}
	\label{abla_number_gaussians}
\end{figure}
\begin{figure}
	\centering 
	\includegraphics[width=\linewidth]{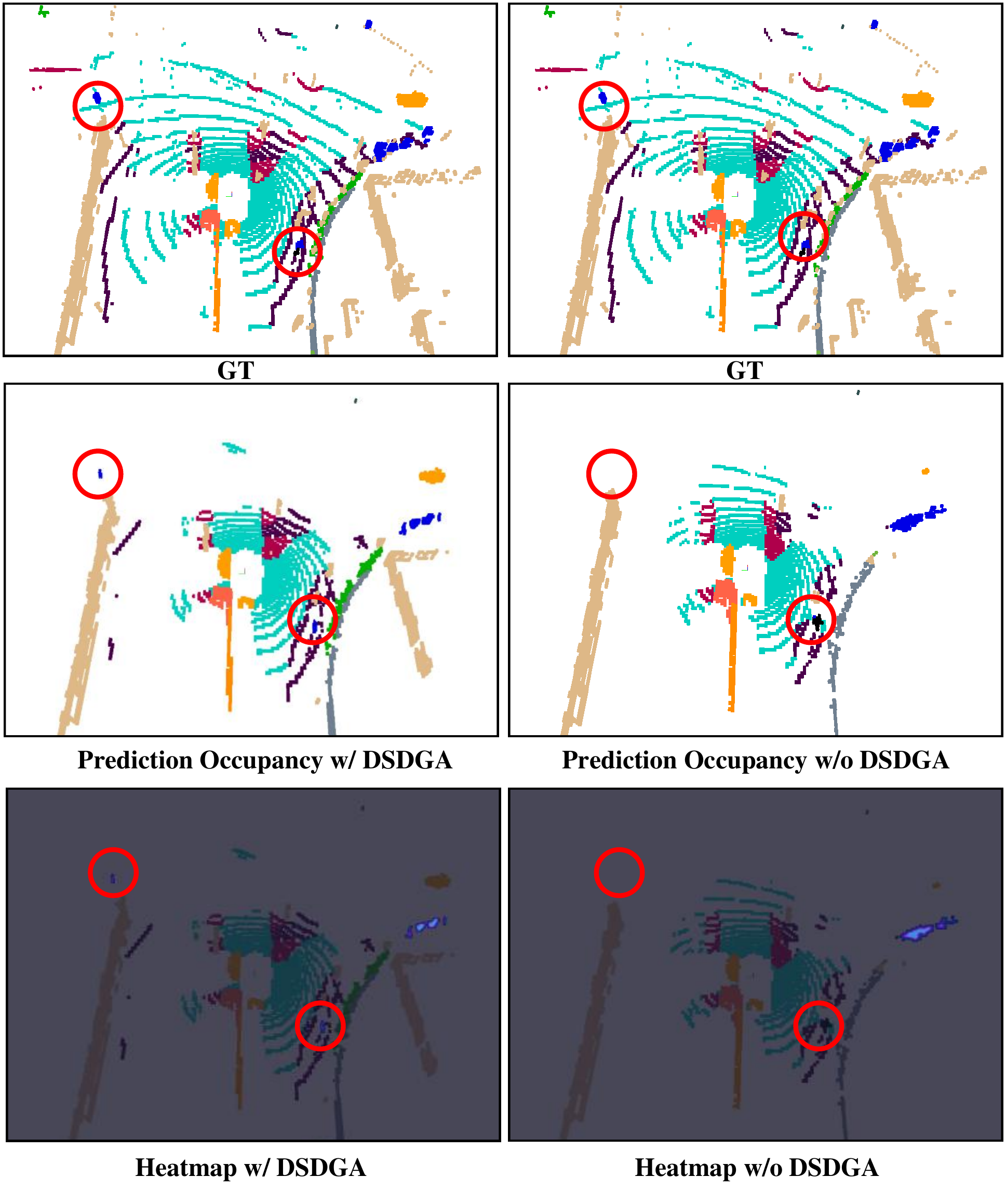}
	\caption{Demonstrating the effect of the DSDGA in the "pedestrian" dynamic objects. The first column is the model with DSDGA module. The second column is the model without DSDGA module. The first row is the GT. The second row is the prediction occupancy. The third row is the heatmap.}
	\label{abla_heatmap2}
\end{figure}
\begin{figure*}
	\centering 
	\includegraphics[width=\linewidth]{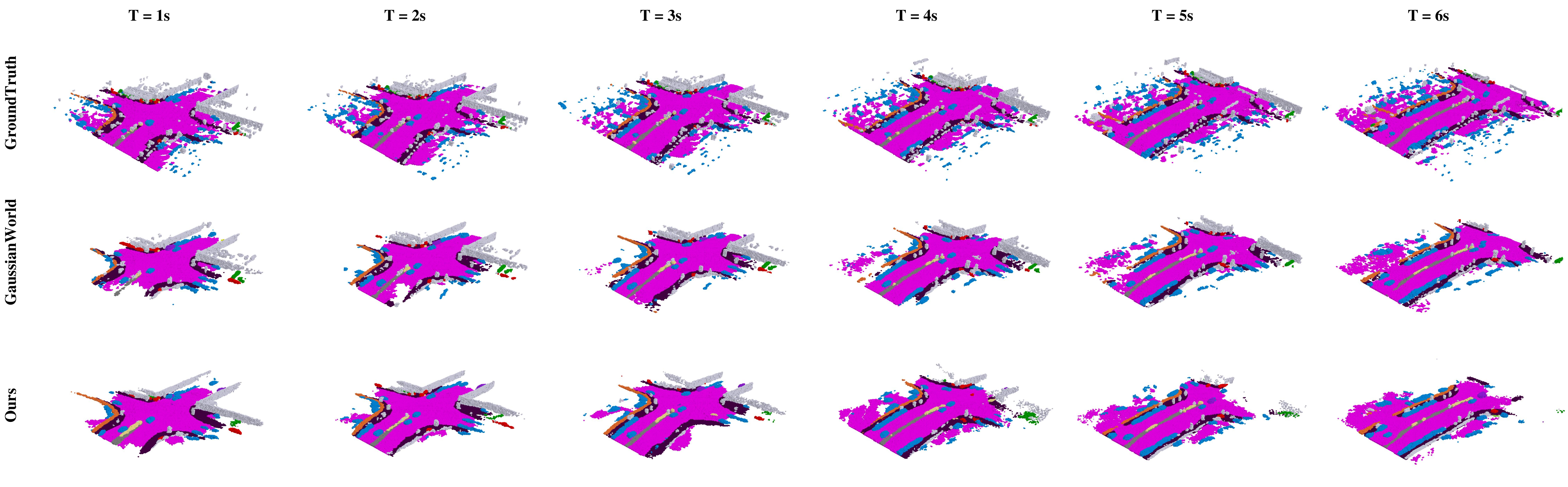}
	\caption{Visualizations of the results of long term occupancy prediction.}
	\label{abla_long_term}
\end{figure*}

\begin{table*}[!t]
	\belowrulesep=0pt
	\aboverulesep=0pt
	\centering     
	\begin{threeparttable}
		\caption{Ablation on the parameters of Top-K and Top-M module without DSDGA.}\label{tbl_abla_k_m}
		\renewcommand{\arraystretch}{1.5}
		\begin{tabular}{c|c|c|c|c|c}
			\toprule
			Methods & Top-K & Top-M & Memory (MB) & mIoU & IoU\\ 
			\hline
			A & [100, 100, 100, 100] & [100, 100, 100, 100] & 3500 & 21. 22 & 33.22 \\
			B & [200, 150, 100, 50] & [200, 150, 100, 50] & 3865 & 21.12& 32.98\\
			C & [50, 25, 10, 5] & [50, 25, 10, 5] & 2010 & 19.51 & 31.21 \\
			D & [100, 75, 50, 20] & [100, 75, 50, 20] & 3320 & 21.31 & 33.41\\
			\bottomrule
		\end{tabular}
	\end{threeparttable}
\end{table*}

\begin{table}[!t]
	\belowrulesep=0pt
	\aboverulesep=0pt
	\centering     
	\begin{threeparttable}
		\caption{Ablation on the DSDGA Module without DGGA and MGA.}\label{tbl_abla_dsdga}
		\renewcommand{\arraystretch}{1.5}
		\begin{tabular}{c|ccc|c}
			\toprule
			\multirow{2}{*}{Methods} & \multicolumn{3}{c|}{mIoU} & \multicolumn{1}{c}{IoU} \\ 
			\cline{2-5}
			& All & Dyna. object & Stat. object & All\\ 
			\hline
			baseline & 18.41 & 16.73 & 20.09 & 29.23 \\
			w/ DCA & 19.85 & 18.99 & 20.71 & 30.93 \\
			w/ SCA & 20.06 & 16.99 & 23.13 & 31.88 \\
			w/ DSDGA  & 22.38 & 19.89 & 24.88 & 32.95 \\
			\bottomrule
		\end{tabular}
	\end{threeparttable}
\end{table}
\subsubsection{Adaptive Fusion}\label{Adaptive Fusion}
In Figs. \ref{fig_af}, we fuse the 3D Gausssian $G_{geo}$ and $G_{sem}$ through the adaptive fusion. In order to fuse $G_{geo}$ and $G_{sem}$ adaptively, the weight map $w=\{w_{n}|n\in N\subseteq\{geo,sem\}\}$ will be generated adaptively. We denote $G={G_{n}|n\in N\subseteq\{geo,sem\}}$. The value of weight map $w_{n}$ can be defined as:
\begin{equation}
	\begin{aligned}
	w_{n}^{i}=\frac{e^{g_{n}^{i}} }{\sum_{j\in N}e^{g_{j}^{i}} },
	\end{aligned}	
\end{equation}
where $g_{n}^{i}$ is the $i$-th gaussians $G_{n}$ and $w_{n}^{i}$ is the $i$-th weight of $w_{n}$.

Finally, we fuse the 3D Gausssian $G_{geo}$ and $G_{sem}$ through the adaptive fusion:
\begin{equation}
	\begin{aligned}
	&G_{out} =  {\textstyle \sum_{n\in N}} G {\textstyle} _{n}\cdot  w_{n} 
	\end{aligned}	
\end{equation}
where $G_{out}$ is the 3D Gaussian after adaptive fusion which is the output of DGGA.
\subsubsection{Multi-scale Graph Attention}\label{MGA}
In this section, we elaborate on the design of a Multi-scale Graph Attention (MGA) mechanism, which is tailored to fuse 3D Gaussians with diverse Top-K and Top-M configurations for hierarchical feature aggregation. As illustrated in Fig. \ref{fig_mga}, the mechanism constructs multi-scale graph structures by defining dynamic neighbor node ranges, specifically setting Top-K as [100, 75, 50, 20] and Top-M as [100, 75, 50, 20]. This strategic parameterization enables the capture of contextual information at varying spatial scales—larger Top values (e.g., 100) facilitate global contextual reasoning, while smaller values (e.g., 20) emphasize local geometric details. Based on these graphs, we generate 3D Gaussians through the DGGA module under different scale graph structures. Finally, we fuse multiple 3D Gaussians using Adaptive Fusion to obtain the multi-scale output 3D Gaussian $G_{multiscale}$.
\subsection{Dynamic-Static Decoupled Gaussian Attention}\label{DSDGA}
In this section, we primarily introduce the Dynamic-Static Decoupled Gaussian Attention (DSDGA) module to further optimize the prediction for dynamic and static objects. This module achieves fine-grained feature separation by leveraging semantic information to distinguish Gaussian anchors into dynamic and static groups, and employs decouple cross attention mechanisms to enable bidirectional feature interaction. Firstly, we decouple dynamic and static Gaussians based on the semantic masks:
\begin{equation}
	\begin{aligned}
	&S = \text{softplus}(G_{:, s:s+d}),\\
	&c_i = \arg\max_{j=1}^{d} S_{ij}, \quad i = 1, 2, \ldots, N,\\
	&m_{\text{static}}[i] = \begin{cases} 
		1, & \text{if}\text{ } c_i \geq t, \\
		0, & \text{otherwise},
		\end{cases}
		\\
	&m_{\text{dynamic}} = \neg m_{\text{static}},\\
	\end{aligned}	
\end{equation}
where $G$ is the 3D Gaussian, $S$ is the semantic score matrix, $c_i$ is the index of the maximum semantic score for the $i$-th Gaussian, $m_{\text{static}}[i]$ is the static mask for the $i$-th Gaussian, $m_{\text{dynamic}}[i]$ is the dynamic mask for the $i$-th Gaussian.
\begin{equation}
	\begin{aligned}
	&G_{\text{dynamic}} = G \odot m_{\text{dynamic}},\\
	&G_{\text{static}} = G \odot m_{\text{static}},\\
	\end{aligned}	
\end{equation}
 where $G_{\text{dynamic}}$ is the dynamic 3D Gaussian, and $G_{\text{static}}$ is the static 3D Gaussian. Subsequently, after obtaining the decoupled dynamic object Gaussians and static scene Gaussians, we employ the proposed Dynamic-Static Decoupled Gaussian Attention (DSDGA) module to interactively optimize the representations of both dynamic objects and static scenes. This module facilitates bidirectional feature interaction through cross-scale attention mechanisms, enabling the exchange of complementary information between the dynamic and static components. Specifically, the DSDGA module consists of two key sub-components:

\subsubsection{Dynamic Cross Attention (DCA)} This sub-module refines the dynamic object representations by leveraging structural and contextual information from the static scene. For instance, it uses prior knowledge of static scene layouts to improve motion prediction of dynamic objects:
\begin{equation}
	\begin{aligned}
	G^{'}_{dynamic}=DCA(Q_{G_{dynamic}},K_{G_{static}},V_{G_{static}}) 
	\end{aligned}	
\end{equation}
\subsubsection{Static Cross Attention (SCA)} This sub-module enhances the static scene representation by selectively attending to dynamic object features that are relevant to scene understanding. For example, it captures how moving agents interact with static environments (e.g., pedestrians on sidewalks, vehicles on roads):
\begin{equation}
	\begin{aligned}
	G^{'}_{static}=SCA(Q_{G_{static}},K_{G^{'}_{dynamic}},V_{G^{'}_{dynamic}}) 
	\end{aligned}	
\end{equation}
Finally, we fuse the refined dynamic and static Gaussian representations $G^{'}_{\text{dynamic}}$ and $G^{'}_{\text{static}}$ through concatenation to generate the final output $G_{\text{final}}$.

\subsection{Other Componets}\label{other}
For other components of the proposed model, we adopt the same structures as those in GaussianFormer and GaussianWorld. Specifically, our model employs ResNet-101 as the image feature extraction module, utilizes 3D sparse convolution \cite{graham2015sparse} for the Self-encoding module, and applies deformable attention \cite{xia2022vision} to fuse image features for optimizing 3D Gaussians. In terms of the temporal alignment module and the temporal fusion approach, we use the stream-based \cite{wang2023exploring, moon2025mitigating,zuo2025gaussianworld} temporal fusion identical to that in GaussianWorld.

\section{Experiments}
In this section, we present the experimental setup and results obtained in this study, structured into three primary subsections for clarity. Section \ref{setting} details the dataset, evaluation metrics and training details employed in our experiments. In Section \ref{comparison sota}, we conduct a comparative analysis of our proposed GraphGSOcc against several state-of-the-art 3D occupancy prediction approaches. Section \ref{ablation study} presents the ablation study, an essential part of our analysis that investigates the contribution of each component within our proposed framework.
\subsection{Setting}\label{setting}
\textbf{Dataset.} We performed occupancy analysis using the SurroundOcc, Occ3D and OpenOcc based on nuScenes dataset, which was captured by a data collection vehicle equipped with a multi-modal sensor array, including one LiDAR, five radars, and six cameras, ensuring full $360^\circ$ environmental coverage. The dataset is partitioned into 700 training and 150 validation scenes, each lasting 20 seconds. High-precision ground truth annotations are provided at 0.5-second intervals, ensuring fine temporal resolution. Spatially, the dataset spans ±40 meters along the x- and y-axes and ranges from -1 to 5.4 meters in the z-axis. Occupancy labels are voxelized at a resolution of $0.4m \times  0.4m \times 0.4m$, covering 17, 17 and 16 semantic categories as defined in SurroundOcc \cite{wei2023surroundocc}, Occ3D \cite{tian2023occ3d} and OpenOcc \cite{graham2015sparse}, respectively. The perceptual data for each driving scene is annotated at 2 Hz, offering a comprehensive and detailed representation for occupancy analysis. 

In addition to the nuScenes dataset, we also evaluate our approach on the KITTI-360 dataset \cite{geiger2012we} to validate the generality of our approach.

\textbf{Metrics.} The semantic segmentation performance is evaluated using both the mean Intersection-over-Union (mIoU) and Intersection-over-Union (IoU):
\begin{equation}
 \begin{aligned}
 &IoU=\frac{TP_{\ne c_{0} } }{TP_{\ne c_{0} }+FP_{\ne c_{0} }+FN_{\ne c_{0} }},\\
 &mIoU=\frac{1}{|C|}\sum_{i\in C} \frac{TP_i}{TP_i+FP_i+FN_i}\\
 \end{aligned}
\end{equation}
where $C$, $c_{0}$, $TP$, $FP$, $FN$ denote the nonempty classes, the empty class, the number of true positive, false positive and false negative predictions, respectively.

\textbf{Training details.}
In our paper, we utilize ResNet-101 as the backbone architecture and feed images with resolutions of $1600 \times 864$ into the model. Our experimental setup is implemented in PyTorch, with a total batch size of 8 distributed over 8 GeForce RTX4090 GPUs. The training epoch is set to 24. We employ the AdamW \cite{loshchilov2017decoupled} optimizer, setting the weight decay to $1\times 10^{-2}$. The learning rate remains constant at $2\times 10^{-4}$, except for a linear warmup during the initial 200 iterations. All loss weights in our methodology are uniformly set to 1.0.

\subsection{Comparison with state-of-the-art (SOTA) methods}\label{comparison sota}
\subsubsection{Quantitative analysis}
To guarantee fair comparisons, all results have been either directly implemented by their respective authors or accurately reproduced by utilizing official codes.

Table \ref{tbl1} presents a comprehensive performance analysis, showcasing how our GraphGSOcc framework compares against the latest advancements in camera-based 3D occupancy prediction on the challenging SurroundOcc dataset. One of the key strengths of GraphGSOcc lies in its modular design, which allows for effortless integration with existing Gaussian-based methodologies. This compatibility enables seamless incorporation into established pipelines without requiring significant architectural modifications.

When benchmarked against previous Gaussian-based techniques, GraphGSOcc demonstrates consistent superiority across multiple performance metrics. By integrating our proposed framework, the resulting model outperforms GaussianFormer by significant margins, achieving a 3.31\% increase in IoU and a 4.38\% improvement in mIoU. Beyond these accuracy gains, the integration also yields notable efficiency enhancements. Specifically, the inference latency is reduced by 92 ms, enabling faster processing times, while GPU memory usage is decreased by a substantial 2109 mb. 

The performance improvements are not limited to GaussianFormer. When integrated with GaussianFormer-2, GraphGSOcc enables the model to achieve the lowest GPU memory consumption of just 2812 MB. As illustrated in Table \ref{tbl1}, the best performance gains are observed when integrating GraphGSOcc into the GaussianWorld framework. These results collectively demonstrate the versatility and effectiveness of our approach across different Gaussian-based architectures, establishing GraphGSOcc as a powerful and practical solution for 3D occupancy prediction tasks.

Table \ref{tbl_open_occ3d} and Table \ref{tbl_kitti} present the performance of our model on the Occ3D, OpenOcc, and KITTI-360 datasets, respectively. Our approach outperforms the previous state-of-the-art methods, demonstrating its strong generalization capability.

\subsubsection{Qualitative analysis}
Our method is compared with high-performance benchmark, GaussianFormer, as illustrated in Figs. \ref{sota} and Figs. \ref{sota2}. In Figs. \ref{sota} and  Figs. \ref{sota2}, we highlight the advantages of our proposed method compared to the other method with red circles. In the first row of Fig. \ref{sota}, the visualization vividly demonstrates the efficacy of our proposed method in predicting the "driveable surface" regions that are otherwise obscured by much "cars". In the second and third rows of Figs. \ref{sota}, GaussianFormer incorrectly predicted "car" and "bus" as "truck," respectively. These results indicate that the proposed method can effectively alleviate the semantic and geometric structure confusion caused by the overlap between different Gaussians. Ultimately, the method we proposed is capable of accurately predicting semantic categories and geometric structures. Figs. \ref{sota2} explicitly demonstrates that our model exhibits superior predictive performance and structural completeness for both dynamic objects and static scenes according to the red circles.

\subsection{Ablation study}\label{ablation study}
In this section, we conduct an ablation study to dissect the contributions of critical components within the GraphGSOcc architecture. We specifically examine the influence of the essential modules: DGGA, MGA, Top-K, Top-M and the number of Gaussians. Through this systematic ablation studies, we evaluate the individual contributions of each module integrated into the GraphGSOcc framework via these ablation studies.

\subsubsection{DGGA module} 
Table \ref{tbl_abla_dgga} demonstrates the effectiveness of our proposed DGGA module, which can improve our proposed method by 2.6\% and 3.1\% in mIoU and IoU, respectively. Further, Table \ref{tbl_abla_dgga} elaborates on the enhanced performance of the refined modules SGA and GGA compared to the original modules in GaussianFormer. Specifically, GGA demonstrates a more pronounced improvement in IoU accuracy, while SGA significantly boosts the mIoU precision.

Figs. \ref{abla_heatmap} demonstrates the accuracy of the proposed SGA module in modeling objects with strong relevance within the same category through interpretable heatmaps. When the SGA module is not adopted, the model incorrectly predicts "bus" as "car". It can be observed from the heatmaps that without the SGA module, the model exhibits prediction ambiguity for similar object categories. In contrast, with the SGA module, the model can more clearly distinguish between similar but different categories of objects.

\subsubsection{MGA module}
For the MGA module, we designed two experiments to verify the number of multi-scale layers and the multi-scale fusion methods. Table \ref{tbl_abla_mga} shows the comparison between three different fusion methods (add, concat, and Conü fusion) and our proposed Adaptive Fusion method. Our Adaptive Fusion can significantly improve the accuracy of mIoU and IoU with a slight increase in GPU memory usage. Figs. \ref{abla_number_dgga} analyzes the impact of the number of multi-scale fusion layers on model performance. As shown in Figs. \ref{abla_number_dgga}, when the number of fusion layers exceeds 4, the prediction accuracy exhibits a downward trend. Therefore, in our proposed model, we adopt 4 as the number of fusion layers for the MGA module.

\subsubsection{The parameters of Top-K and Top-M} 
As key parameters for SGA and GGA, Top-K and Top-M play a critical role: too few graph nodes are insufficient to model numerous objects in 3D scenes, while excessive graph nodes may cause irrelevant objects to interfere with each other. Therefore, Table \ref{tbl_abla_k_m} analyzes the performance of our model under different numbers of graph nodes. When both Top-K and Top-M are set to [100, 75, 50, 20].
\subsubsection{DSDGA module}
Table \ref{tbl_abla_dsdga} demonstrates the effectiveness of the proposed DSDGA module. When this module is employed, the model exhibits better prediction performance for both moving objects and static scenes as a whole. The integration of the SCA leads to a significant improvement in the model's predictive performance for static long-range contexts, while the adoption of the DCA results in a notable enhancement in the prediction of dynamic objects. This further validates that the decoupled interactive attention mechanism for dynamic and static objects can effectively facilitate the prediction of dynamic objects and static scenes. 

Figs \ref{abla_heatmap2} presents the heatmap verifying the effectiveness of the DSDGA module. It can be observed from the heatmap that after incorporating DSDGA, the model demonstrates effective prediction for dynamic pedestrians, and successfully focuses features on moving pedestrians.
\subsubsection{The number of Gaussians} 
Figs. \ref{abla_number_gaussians} analyzes the performance of our model under different numbers of Gaussians. As shown in Figs. \ref{abla_number_gaussians}, our model demonstrates superior performance with fewer Gaussians: it exceeds the prediction performance of GaussianFormer with 144,000 Gaussians even when only 5,000 Gaussians are used. By requiring fewer Gaussians, our model consumes less GPU memory, thereby enhancing model efficiency.

\subsection{Visualization on Long Time prediction}
Figs \ref{abla_long_term} systematically evaluates the long-sequence occupancy prediction capabilities of GraphGSOcc against GaussianWorld, showcasing distinct advantages in temporal consistency and geometric stability. Qualitative analyses reveal that our model maintains drivable area continuity over 6-second sequences compared with GaussianWorld. Visually, our method can predict drivable areas more continuously, and the geometric shapes in the long-term prediction changes of objects are more stable compared with GaussianWorld. The adaptive graph attention mechanisms enable consistent modeling of topological relationships, reducing semantic misclassification errors in long-term sequences. These findings highlight GraphGSOcc's superiority in handling temporal dynamics for real-world autonomous driving scenarios.

\section{Conclusion}
In conclusion, our proposed GraphGSOcc model marks a significant advancement in the realm of 3D semantic occupancy prediction within the context of vision-centric autonomous driving perception. By leveraging the potential of 3D Gaussian Splatting (3DGS), GraphGSOcc effectively addresses the limitations of existing 3DGS-based methods, offering a more efficient and accurate solution for capturing the complex 3D structure of real-world scenes.

The Dual Gaussian Graph Attention (DGGA) mechanism and the Multi-scale Graph Attention (MGA) framework, as the core components of GraphGSOcc, play a crucial role in enhancing the model's performance. The dynamic construction of geometric and semantic graphs through DGGA enables the model to adaptively optimize both boundary details and object-level topology, which effectively mitigates position drift and semantic ambiguities at object boundaries. Meanwhile, the MGA framework refines Gaussians hierarchically, optimizing both boundary details and small objects, which effectively mitigates position drift and small objects loss. Additionally, the Dynamic-Static Decoupled Gaussian Attention (DSDGA) mechanism decouples dynamic and static objects, enabling the model to optimize the prediction for dynamic and static objects separately.

Our experimental results demonstrate the superiority of GraphGSOcc. Achieving a mIoU of 25.20\% on Surroundocc and reducing GPU memory usage to 6.8 GB, our method not only improves the accuracy of 3D semantic occupancy prediction but also significantly enhances computational efficiency. 

Future research could focus on integrating more advanced temporal information to handle dynamic scenes more effectively, as well as exploring the application of GraphGSOcc in more diverse driving scenarios.

\bibliographystyle{unsrt}  


\end{document}